\documentclass[11pt]{article}
\usepackage{graphicx}
\usepackage{amsmath,amssymb}
\usepackage{bm}

\usepackage{booktabs}
\usepackage{algorithm}
\usepackage{algpseudocode}

\usepackage{abstract}
\usepackage[title]{appendix}

\allowdisplaybreaks[4]

\graphicspath{{figures/}}

\makeatletter
\DeclareMathOperator\softplus{sfp}
\DeclareMathOperator\sigmoid{sig}
\DeclareMathOperator\logit{logit}
\makeatother

\begin{document}
\begin{center}
\huge{\textsc{Effective Method for Inverse Ising Problem under Missing Observations in Restricted Boltzmann Machines}} \\
\vspace{5mm}
\large{Kaiji Sekimoto\footnote{k.sekimoto1002@gmail.com} and Muneki Yasuda} \\
\vspace{2mm}
\normalsize{Graduate School of Science and Engineering, Yamagata University, Japan}
\end{center}

\vspace{2mm}
\begin{abstract}
\vspace{-2.5mm}
Restricted Boltzmann machines (RBMs) are energy-based models analogous to the Ising model and are widely applied in statistical machine learning. 
The standard inverse Ising problem with a complete dataset requires computing both data and model expectations and is computationally challenging because model expectations have a combinatorial explosion.
Furthermore, in many applications, the available datasets are partially incomplete, making it difficult to compute even data expectations.
In this study, we propose a approximation framework for these expectations in the practical inverse Ising problems that integrates mean-field approximation or persistent contrastive divergence to generate refined initial points and spatial Monte Carlo integration to enhance estimator accuracy. 
We demonstrate that the proposed method effectively and accurately tunes the model parameters in comparison to the conventional method.
\end{abstract}

\section{Introduction} \label{sec:Introduction} 

The restricted Boltzmann machine (RBM)~\cite{smolensky1986,hinton2002}, a variant of the Ising model, is widely used in statistical machine learning and has been applied to a variety of tasks, including retraining deep neural networks~\cite{salakhutdinov2010}, collaborative filtering~\cite{salakhutdinov2007}, anomaly detection~\cite{do2018energy}, time-series analysis~\cite{kuremoto2014}, automated molecular design~\cite{Hatakeyama-Sato2022}, and quantum state tomography and wave function construction for quantum many-body systems~\cite{torlai2018,carleo2017}.
The inverse Ising problem, known as model training in the field of machine learning, aims to estimate parameters such that the Ising model fits the data distribution.
This is typically performed by maximizing the likelihood of a complete dataset using the gradient ascent method.
The gradients with respect to the parameters are given by the difference between the expectation for the data distribution and the expectation for the model distribution, referred to as clamped and free expectations, respectively.
Although the clamped expectation can be evaluated analytically, the free expectation cannot be computed directly because it requires intractable multiple summations over all possible configurations of random variables.
Thus, solving the standard inverse Ising problem relies on approximations for the free expectations.
Because it is empirically known that the performance of likelihood maximization depends on the accuracy of these approximations~\cite{yasuda2018,Sekimoto2023}, a high-accuracy approximation is desirable.

The standard inverse Ising problem requires a complete dataset.
Nevertheless, in practice some elements of each data point may occasionally be unobservable owing to issues such as monitoring device failures, database errors, or noise in data transmission lines.
Although this can be addressed by restoring an incomplete dataset to a complete one through removal or imputation, these approaches would reduce the available data or introduce unintended noise.
In RBM, a method for solving the inverse Ising problem without restoration has been proposed~\cite{fissore2019robust}.
This approach introduces a likelihood function that marginalizes the missing variables for each incomplete data point.
The gradients for this likelihood have the same form as in the standard inverse Ising problem.
However, the marginalization over missing variables makes it difficult to evaluate not only the free expectations but also the clamped expectations computationally.
Therefore, solving the inverse Ising problem with an incomplete dataset requires approximations for both expectations.

The conventional method, known as lossy contrastive divergence (Lossy-CD)~\cite{fissore2019robust} evaluates both expectations using a sampling approximation based on blocked Gibbs sampling (bGS) and Monte Carlo integration (MCI).
The bGS is a Markov chain Monte Carlo (MCMC) method that be easily applied to RBMs.
In Lossy-CD, sample points are drawn from the target distribution using bGS where initial points are uniformly generated, and using the sample points, the target expectation is approximated by MCI, i.e., a sample average.
For a highly accurate sampling approximation, it is essential to generate samples that follow the model distribution and an efficient estimator.
To achieve this with Lossy-CD, we require a long MCMC step (i.e., a long burn-in time) and a lager number of sample points.
This implies that the high-performance inverse Ising problem based on Lossy-CD is computationally expensive.
In this study, we improve both the sampling and approximation methods of Lossy-CD to accurately solve the inverse Ising problem without increasing computational costs.
Specifically, we aim to obtain initial points for bGS that closely follow the target distribution and to develop a more efficient estiamtor.
We approximate the free expectations using the sampling procedure of persistent contrastive divergence (PCD)~\cite{tieleman2008training} and spatial MCI (SMCI)~\cite{yasuda2015smci,yasuda2021smci}.
PCD provides a sampling approximation for the free expectation required in the standard inverse Ising problem in RBMs and sets initial points to the sample points used in the previous parameter update, allowing for high-quality sampling with only a few MCMC steps.
SMCI is an extension of MCI that explicitly considers the graph structure of the Ising model and provides an estimator with a lower asymptotic variance than MCI by partially performing the summation.
We approximate the clamped expectations using mean-field approximation (MFA)~\cite{doi:10.7566/JPSJ.85.034001,decelle2021restricted} and SMCI.
MFA approximates the target distribution using a test distribution, which is defined as the product of individual distributions for each variable and is obtained by minimizing the Kullback-Leibler divergence (KLD) between the test and target distributions under the normalization constraint of the test distribution.
The minimization is performed by solving self-consistent mean-field equations, which generally have multiple solutions.
To obtain multiple test distributions, we first prepare random values for the desired number of sample points.
For each value, we independently solve the mean-field equation using the successive substitution method.
These test distributions are then used to generate a set of initial points.
Since the test distributions are obtained by minimizing the KLD, the initial points generated from them are expected to follow the target distribution more closely than randomly generated initial points used in Lossy-CD.
Notably, the strategy of using MFA to initialize sampling has also been employed to approximate the partition function via annealed importance sampling~\cite{neal2001}, demonstrating its effectiveness~\cite{prat2025mean}. 
In this study, we adopt this strategy for the inverse Ising problem.
By providing high-quality initial points and efficient estimators, the proposed method could maximize the likelihood more accurately than Lossy-CD.

The remainder of this paper is organized as follows. 
Section~\ref{sec:rbm} reviews the basic formulation of RBM.
Section~\ref{sec:inverse_Ising_prob} describes the inverse Ising problem with incomplete datasets and summarizes the existing Lossy-CD method. 
Section~\ref{sec:base_method} introduces the base methods for improving approximation accuracy, including MFA (see Section~\ref{sec:mfa}) and SMCI (see Section~\ref{sec:smci}). Section~\ref{sec:improvement} presents our proposed improvements in sampling and estimation for both clamped and free expectations. 
Numerical experiments that validate the proposed method are reported in Section~\ref{sec:num_exp}. 
Finally, Section~\ref{sec:conclusion} concludes the paper and outlines directions for future work.

\section{Restricted Boltzmann Machines} \label{sec:rbm}

An RBM is an MRF defined on a complete bipartite graph and consists of two layers: a visible and a hidden layers.
The visible layer contains visible variables $\bm{v}:=\{v_i \in\{0,1\}\mid i\in V\}$, which correspond to a single data point, and the hidden layer contains hidden variables $\bm{h}:=\{h_j \in\{0,1\}\mid j\in H\}$, which control the expression power of the RBM.
Here, $V:=\{1,2,\ldots,n\}$ and $H:=\{n+1,n+2,\ldots,n+m\}$ denotes the sets of node labels in the visible and hidden layers, respectively.
We denote the visible variables corresponding to a subset $A\subseteq V$ as $\bm{v}_A = \{v_i\mid i\in A\}$.
The energy function for the visible and hidden variables is defined as
\begin{align}
E_\theta(\bm{v},\bm{h}) := - \sum_{i\in V}b_iv_i - \sum_{j\in H}c_jh_j - \sum_{i\in V}\sum_{j\in H}w_{i,j}v_ih_j,
\label{eq:RBM_Energy}
\end{align}
where $b_i, c_j$ are bias parameters assigned to a visible node $i$ and a hidden node $j$, respectively, and $w_{i,j}$ is a connection parameter assigned to the link between visible $i$ and hidden $j$ nodes.
We collectively denote the parameters $b_i, c_j$ and  $w_{i,j}$ by $\theta$.
Using the energy function in Eq.~\eqref{eq:RBM_Energy}, the joint distribution over $\bm{v}$ and $\bm{h}$ is defined as
\begin{align}
P_\theta(\bm{v},\bm{h}) := \frac{1}{Z_\theta}\exp(-E_{\theta}(\bm{v},\bm{h})),
\label{eq:RBM_Dist}
\end{align}
where $Z_\theta$ is the partition function (or normalizing constant), given by
\begin{align}
Z_\theta := \sum_{\bm{v}}\sum_{\bm{h}}\exp(-E_{\theta}(\bm{v},\bm{h})),
\label{eq:RBM_Part-Func}
\end{align}
where $\sum_{\bm{v}} := \prod_{i\in V}\sum_{v_i\in\{0,1\}}$ is a multiple summation over $\bm{v}$ and $\sum_{\bm{h}}:=\prod_{j\in H} \sum_{h_j\in\{0,1\}}$ is that over $\bm{h}$.
The conditional distributions for one layer given the other layer are expressed as
\begin{align}
P_\theta(\bm{v}\mid\bm{h}) &= \prod_{i \in V} \mathrm{Ber}\bigl(v_i\mid\sigmoid(\lambda_i(\bm{h}))\bigr), \label{eq:cond_v|h} \\
P_\theta(\bm{h}\mid\bm{v}) &= \prod_{j\in H} \mathrm{Ber}\bigl(h_j\mid\sigmoid(\tau_j(\bm{v}))\bigr), \label{eq:cond_h|v}
\end{align}
where $\mathrm{Ber}(x\mid \rho)$ denotes the Bernoulli distribution with parameter $\rho \in (0,1)$, representing the probability of $x = 1$.
Here, $\sigmoid(x):=1/(1+e^{-x})$ is the sigmoid function, and
\begin{align*}
\lambda_i(\bm{h}) &:= b_i + \sum_{j\in H} w_{i,j} h_j, \\
\tau_j(\bm{v}) &:= c_j + \sum_{i\in V} w_{i,j} v_i.
\end{align*}
The conditional distributions in Eqs.~\eqref{eq:cond_v|h} and \eqref{eq:cond_h|v} show that when the variables in one layer are fixed, the variables in the other layer are statistically independent. 
This property is referred to as conditional independence and facilitates an efficient MCMC method named bGS shown in Algorithm~\ref{alg:blocked_Gibbs_sampling}.
For high-quality sampling, we require initial points that are close to the target distribution, or alternatively a larger MCMC step to break the correlation with initial points that do not follow the target distribution.

\begin{algorithm}[t]
\caption{Blocked Gibbs sampling on RBMs}
\label{alg:blocked_Gibbs_sampling}
\begin{algorithmic}[1]
\Require a sample size $K$, a MCMC step $R$, initial points for the visible variables $\{\mathbf{v}_{0}^{(\nu)}\in \{0,1\}^n\mid\nu=1,2,\ldots,K\}$
\For{$\nu=1,2,\ldots,K$}
\State $\mathbf{h}_{0}^{(\nu)}\sim P_\theta(\bm{h}\mid\mathbf{v}_{0}^{(\nu)})$
\For{$r=1,2,\ldots,R$}
\State $\mathbf{v}_{r}^{(\nu)}\sim P_\theta(\bm{v}\mid\mathbf{h}_{r-1}^{(\nu)})$
\State $\mathbf{h}_{r}^{(\nu)}\sim P_\theta(\bm{h}\mid\mathbf{v}_{r}^{(\nu)})$
\EndFor
\EndFor
\Ensure $K$ sample points $\bigl\{\{\mathbf{v}_{R}^{(\nu)},\mathbf{h}_{R}^{(\nu)}\}\mid\nu=1,2,\ldots,K\bigr\}$
\end{algorithmic} 
\end{algorithm}

\section{Inverse Ising Problem using Incomplete Dataset in Restricted Boltzmann Machines} \label{sec:inverse_Ising_prob}

Suppose we obtain an incomplete dataset consisting of $N$ missing data points, expressed as
\begin{align*}
\mathcal{D} &:= \{\mathbf{d}^{(\mu)}\mid \mu=1,2,\ldots,N\}, \\
\mathbf{d}^{(\mu)} &:= \{\mathrm{d}_i^{(\mu)}\in\{0,1\}\mid i\in O^{(\mu)}\subseteq V\},
\end{align*}
where $O^{(\mu)}$ denotes the observed region of the $\mu$th data point, and the corresponding missing region is denoted by $M^{(\mu)}:=V \backslash O^{(\mu)}$.
We assume the mechanism of missing data is ignorable; that is, the missing is completely at random, or the missing is at random and the parameter of the mechanism is distinct from $\theta$.
From this assumption, the inverse Ising problem of RBMs using the incomplete dataset is performed by maximizing the log-likelihood, defined as
\begin{align}
\ell_{\mathcal{D}}(\theta) := \frac{1}{N}\sum_{\mu=1}^N \ln \left[\sum_{\bm{v}_{M^{(\mu)}}}\sum_{\bm{h}}P_\theta(\mathbf{d}^{(\mu)}, \bm{v}_{M^{(\mu)}}, \bm{h})\right].
\label{eq:log-likelihood}
\end{align}
The maximization is performed using the gradient ascent method, with the gradients of the log-likelihood with respect to the parameters $\theta$.
The gradients are expressed as follow:
\begin{align}
\frac{\partial\,\ell_{\mathcal{D}}}{\partial\,b_i} &= \frac{1}{N}\sum_{\mu=1}^N \Bigl[I_{O^{(\mu)}}(i)\, \mathrm{d}_i^{(\mu)} + (1 - I_{O^{(\mu)}}(i)) \, \mathbb{E}_\theta[v_i\mid\mathbf{d}^{(\mu)}]  \Bigr] - \mathbb{E}_\theta[v_i] \label{eq:grad_b}\\ 
\frac{\partial\,\ell_{\mathcal{D}}}{\partial\,c_j} &= \frac{1}{N}\sum_{\mu=1}^N \mathbb{E}_\theta[h_j\mid\mathbf{d}^{(\mu)}] - \mathbb{E}_\theta[h_j] \label{eq:grad_c} \\
\frac{\partial\,\ell_{\mathcal{D}}}{\partial\,w_{i,j}} &= \frac{1}{N}\sum_{\mu=1}^N \Bigl[I_{O^{(\mu)}}(i)\,  \mathrm{d}_i^{(\mu)}\,\mathbb{E}_\theta[h_j\mid\mathbf{d}^{(\mu)}] + (1 - I_{O^{(\mu)}}(i)) \, \mathbb{E}_\theta[v_ih_j\mid\mathbf{d}^{(\mu)}]\Bigr] - \mathbb{E}_\theta[v_i h_j] \label{eq:grad_w} 
\end{align}
where $I_{O^{(\mu)}}(i)$ is the indicator function that returns $1$ if $i \in O^{(\mu)}$ and return $0$ if $i \notin O^{(\mu)}$.
The first term in the gradients contains the conditional expectation given a data point, defined by
\begin{align}
\mathbb{E}_\theta[f(\bm{v}_{M^{(\mu)}},\bm{h})\mid\mathbf{d}^{(\mu)}] &:= \sum_{\bm{v}_{M^{(\mu)}}} \sum_{\bm{h}} f(\bm{v}_{M^{(\mu)}},\bm{h}) P_{\theta}(\bm{v}_{M^{(\mu)}}, \bm{h}\mid\mathbf{d}^{(\mu)}), 
\label{eq:clamped_expect}
\end{align}
and the second term is the standard expectation defined by
\begin{align}
\mathbb{E}_\theta[f(\bm{v},\bm{h})] := \sum_{\bm{v}}\sum_{\bm{h}} f(\bm{v},\bm{h}) P_{\theta}(\bm{v},\bm{h}). 
\label{eq:free_expect}
\end{align}
We henceforth refer to the expectations in Eqs.~\eqref{eq:clamped_expect} and \eqref{eq:free_expect} as clamped and free expectations, respectively, and the conditional distribution $P_{\theta}(\bm{v}_{M^{(\mu)}}, \bm{h}\mid\mathbf{d}^{(\mu)})$ as the clamped distribution.
Because these expectations involve multiple summations that are computationally intractable, approximations are required in general.
A conventional method, called Lossy-CD~\cite{fissore2019robust}, approximates the expectations using a sampling-based approach, which employs bGS with initial points independently generated from a uniform distribution, and MCI.
For example, when to approximate the free expectation, we generate a sample set, $\bigl\{\bigl\{\mathbf{v}^{(\nu)},\mathbf{h}^{(\nu)}\bigr\}\mid\nu=1,2,\ldots,K\bigr\}$, from the distribution in Eq.~\eqref{eq:RBM_Dist} using bGS, and approximate the free expectation by MCI as follows:
\begin{align*}
\mathbb{E}_\theta[f(\bm{v},\bm{h})] \approx \frac{1}{K}\sum_{\nu=1}^{K} f\bigl(\mathbf{v}^{(\nu)},\mathbf{h}^{(\nu)}\bigr).
\end{align*}
The computational cost of this approximation is $O(KRnm)$ because the computational cost of sampling by bGS with a sample size $K$ and an MCMC step $R$ is $O(KRnm)$, and the cost of estimation by MCI is $O(K)$.
We note that the computational bottleneck in Lossy-CD is the sampling step by bGS.

It is known the performance of maximizing the log-likelihood depends on the approximation accuracy of the expectations~\cite{yasuda2018,Sekimoto2023}; therefore, a high-accuracy approximation is desired.
To achieve this with Lossy-CD, a large number of MCMC steps is required to eliminate correlations with the initial points and ensure sampling from the target distribution.
Additionally, a large number of sample points is needed, as suggested by the central limit theorem.
This implies that high-performance inverse Ising problems based on Lossy-CD are computationally expensive.
In this study, we improve both the sampling and approximation methods of Lossy-CD to accurately solve the inverse Ising problem without increasing computational costs.

\section{Underlying Methods for Approximation Improvement} \label{sec:base_method}

In this section, we introduce MFA and SMCI that are underlying bases of improving Lossy-CD.
To explain the underlying methods, we consider a Boltzmann machine (BM) defined on an undirected graph $G(\Omega, E)$ expressed as
\begin{align}
P_{\mathrm{BM}}(\bm{x}) \propto \exp\left(\sum_{i\in\Omega}h_ix_i + \sum_{(i,j)\in E} J_{i,j}x_ix_j\right),
\label{eq:bm}
\end{align}
where $\Omega$ is the set of node labels, and $E$ is the set of undirected edges where the edges $(i,j)$ and $(j,i)$ represent the same edge.
Here, $\bm{x}:=\{x_i\in\{0,1\}\mid i\in\Omega\}$ denotes the random variables, and $h_i$ and $J_{i,j}$ are the bias and the symmetric interaction (i.e., $J_{i,j} = J_{j,i}$), respectively.
An RBM in Eq.~\eqref{eq:RBM_Dist} is a special case of BMs, where $\Omega=V\cup H$ and $E = V\times H$.
We denote the variables for a region $A\subseteq\Omega$ as $\bm{x}_A :=\{x_i\mid i\in A\}$, and the multiple summation over $\bm{x}_A$ as $\sum_{\bm{x}_A}=\prod_{i\in A}\sum_{x_i\in\{0,1\}}$.
Additionally, we denote the adjacency set of a region $A\subseteq\Omega$ as $\partial A := \{j\mid (i,j)\in E, i\in A, j\notin A\}$.

\subsection{Mean Field Approximation} \label{sec:mfa}

MFA approximates the target distribution in Eq.~\eqref{eq:bm} by a test distribution defined by
\begin{align}
Q(\bm{x}) := \prod_{i\in \Omega}Q_i(x_i),
\label{eq:MFA_dist}
\end{align}
where $Q_i(x_i)$ denotes the marginal test distribution for $x_i$, $Q_i(x_i) = \sum_{\bm{x}\backslash\{x_i\}} Q(\bm{x})$.
The form of the test distribution is obtained by variational minimization of the Kullback--Leibler divergence (KLD) between the test and target distributions,
\begin{align*}
\mathrm{KL}(Q\,||\,P_{\mathrm{BM}}) := \sum_{\bm{x}} Q(\bm{x})\ln\frac{Q(\bm{x})}{P_\mathrm{BM}(\bm{x})},
\end{align*}
with respect to $Q_i$ under the normalizing constraints $\sum_{x_i\in\{0,1\}} Q_i(x_i) = 1$ for $i\in \Omega$.
The minimization can be analytically performed by the method of Lagrange multipliers, and the marginal test distribution then results as
\begin{align}
Q_i(x_i) = \mathrm{Ber}\bigl(x_i\mid \sigmoid(\alpha_i(\bm{m}_{\partial \{i\}}))\bigr)\qquad(i\in\Omega),
\label{eq:MFA}
\end{align}
where $\alpha_i(\bm{m}_{\partial \{i\}}) := h_i + \sum_{j\in \partial \{i\}}J_{i,j} m_j$, and $\bm{m}:=\{m_i \mid i\in \Omega\}$ are obtained by solving self-inconsistent equations called the mean-field equations:
\begin{align*}
m_i = \sigmoid(\alpha_i(\bm{m}_{\partial \{i\}}))\qquad(i\in \Omega).
\end{align*}
The resulting test distribution in Eq.~\eqref{eq:MFA} has a tractable form, allowing for easy sampling.

From the design of the test distribution in Eq.~\eqref{eq:MFA_dist}, MFA can provide an accurate approximation distribution when the interactions are negligible (i.e., their magnitudes are close to zero or significantly smaller than those of the biases).
In other words, the test distribution can accurately capture the entire target distribution when the target distribution has a smooth energy landscape or is unimodal. 
However, when the interactions are not negligible (i.e., the target distribution becomes multimodal), the test distribution cannot capture the multimodality of the target distribution and tends to reproduce only a local peak.

\subsection{Spatial Monte Carlo Integration} \label{sec:smci}

SMCI is a spatial extension of the standard MCI.
Given sample points, standard MCI approximates an expectation by the sample average.
On the other hand, SMCI leverages the graph structure of the Ising model and approximates an expectation using the sample average of the conditional expectation of the target variables.
In this sense, SMCI can be interpreted as a method that applies Rao--Blackwellization~\cite{liu2001} to the approximation of an expectation in the Ising model.
Let us approximate an expectation for a function $f(\bm{x}_T)$ is defined by
\begin{align}
\mathbb{E}_{\mathrm{BM}}[f(\bm{x}_T)] := \sum_{\bm{x}} f(\bm{x}_T) P_{\mathrm{BM}}(\bm{x}),
\label{eq:bm_expectation}
\end{align}
where $T\subseteq \Omega$ is the target region.
Suppose that we obtain $M$ sample points, $\{\mathbf{x}^{(\nu)}\mid\nu=1,2,\ldots,K\}$, generated from $P_{\mathrm{BM}}(\bm{x})$.
In standard MCI, the expectation is approximated by
\begin{align}
\mathbb{E}_{\mathrm{BM}}[f(\bm{x}_T)] \approx \frac{1}{K}\sum_{\nu=1}^{K} f(\mathbf{x}_T^{(\nu)}),
\label{eq:mci}
\end{align}
where $\mathbf{x}_T^{(\nu)}$ denotes the $\nu$-th sample point of $\bm{x}_T$.
In SMCI, we introduce a sum region $U$, which includes the target region (i.e., $T\subseteq U\subseteq \Omega$), and consider the conditional distribution of $\bm{x}_U$ expressed as
\begin{align}
P_{\mathrm{BM}}(\bm{x}_U\mid\bm{x}_{\Omega\backslash U}) = P_{\mathrm{BM}}(\bm{x}_U\mid\bm{x}_{\partial U}).
\label{eq:spatial_markov_property}
\end{align}
The equation is derived by utilizing the spatial Markov property of MRFs.
Using the conditional distribution in Eq.~\eqref{eq:spatial_markov_property}, the expectation in Eq.~\eqref{eq:bm_expectation} is approximated by
\begin{align}
\mathbb{E}_{\mathrm{BM}}[f(\bm{x}_T)] \approx \frac{1}{K}\sum_{\nu=1}^{K} f_{(T,U)}(\mathbf{x}_{\partial U}^{(\nu)}),
\label{eq:smci}
\end{align}
where $f_{(T,U)}(\mathbf{x}_{\partial U}^{(\nu)})$ denotes the conditional expectation of the function $f(\bm{x}_{T})$ defined by
\begin{align*}
f_{(T,U)}(\bm{x}_{\partial U}) := \sum_{\bm{x}_U} f(\bm{x}_T) P_{\mathrm{BM}}(\bm{x}_U\mid\bm{x}_{\partial U}).
\end{align*}
The summation over the variables for the sum region $U$, $\sum_{\bm{x}_{U}}$ is analytically or numerically performed.

The estimators of the standard MCI and SMCI in Eqs.~\eqref{eq:mci} and \eqref{eq:smci} are both unbiased, consistent, and asymptotically normal~\cite{yasuda2021smci}.
Furthermore, the two properties for the asymptotic variance have been proven~\cite{yasuda2015smci,yasuda2021smci}: for a given sample set, (i) the asymptotic variance of the SMCI estimator is smaller than that of the standard MCI estimator, and (ii) the asymptotic variance of the SMCI estimator with the sum region $U_1$ is smaller than that with the sum region $U_2$ when $U_1\supset U_2$.
Although property (ii) states that the approximation accuracy improves as the sum region expands, the expansion generally causes an exponential increase in the computational cost because the cost of Eq.~\eqref{eq:smci} is $O(KF2^{|U|})$, where the computational cost of $f(\bm{x}_T)$ is $O(F)$.
Therefore, we should expand the sum region within the limits of the computational cost, except in special cases, e.g., when the sum region is a planar or a tree graph~\cite{yasuda2021smci}.

\begin{figure}
    \centering
    \includegraphics[width=0.9\linewidth]{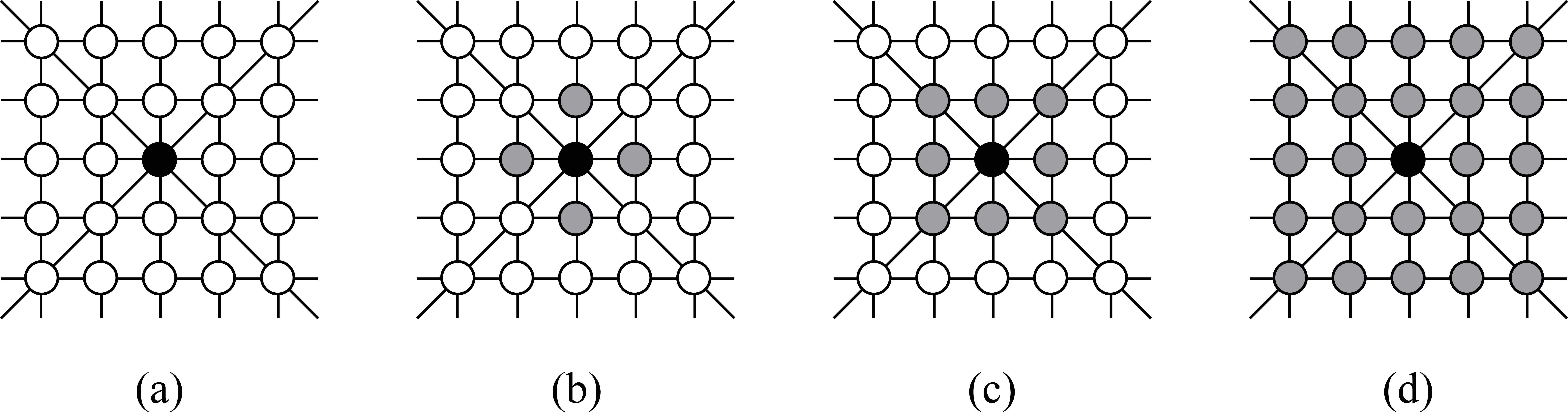}
    \caption{Illustrations of sum regions in (a) 1-SMCI, (b) an example of s2-SMCI, (c) 2-SMCI, and (d) 3-SMCI. The target region is shown in black, and each sum region corresponds to the union of the target region and the region highlighted in gray.}
    \label{fig:smci}
\end{figure}

Since the sum region can be freely selected, several selection strategies have been provided.
The SMCI with a sum region that includes the $(k-1)$th nearest neighboring region of the target region is called the $k$th-order SMCI ($k$-SMCI).
In particular, when a sum region is identical to the target region (i.e, $U=T$), the SMCI corresponds to 1-SMCI.
Moreover, as an intermediate approach between 1- and 2-SMCIs, the s2-SMCI is introduced.
The sum region of the s2-SMCI consists of the target region and an independent subset of the adjacent region of the target region, i.e.,
\begin{align*}
U = T\cup \{i\mid i\in\partial T, \forall j\in \partial T, (i,j)\notin E\}.
\end{align*}
Because the summation over the variables for the independent subset can be analytically computed, the computational cost of s2-SMCI is equivalent to that of 1-SMCI.
From the above discussion about the asymptotically variance, the approximation accuracy increases as $k$ increases, and the accuracy of the s2-SMCI is between 1- and 2-SMCI.
Examples of sum regions in $k$-SMCI and s2-SMCI are illustrated in Figure~\ref{fig:smci}.

\section{Improving Sampling Approximation for Clamped and Free Expectations} \label{sec:improvement}

In this section, we propose a method for solving inverse Ising problems by approximating the clamped expectation based on MFA and SMCI, and the free expectation based on PCD and SMCI.

\subsection{Improvement on Sampling} \label{sec:improvement_sampling}

The clamped expectation in Eq.~\eqref{eq:clamped_expect} is computed for each data point $\mathbf{d}^{(\mu)}$ and must be approximated with high accuracy within a limited number of MCMC steps.
To achieve this, bGS requires the use of initial points that closely follow the model distribution.
Therefore, we propose a method for generating initial points based on MFA.
Let us consider a test distribution corresponding to the clamped distribution expressed as
\begin{align}
Q(\bm{v}_{M^{(\mu)}},\bm{h}\mid\mathbf{d}^{(\mu)}) := \Biggl(\prod_{i\in M^{(\mu)}} Q_i(v_i\mid\mathbf{d}^{(\mu)})\Biggr)\Biggl(\prod_{j\in H} Q_j(h_j\mid\mathbf{d}^{(\mu)})\Biggr),
\label{eq:test_distribution}
\end{align}
where $Q_i(v_i\mid\mathbf{d}^{(\mu)})$ and $Q_j(h_j\mid\mathbf{d}^{(\mu)})$ denote the marginal test distribution for $v_i$ and $h_j$, respectively.
From Eq.~\eqref{eq:MFA}, the marginal distributions are derived as
\begin{align*}
Q_i(v_i\mid\mathbf{d}^{(\mu)}) &= \mathrm{Ber}\bigl(v_i\mid\sigmoid(\lambda_i(\bm{m}_h^{(\mu)}))\bigr),  \\
Q_j(h_j\mid\mathbf{d}^{(\mu)}) &= \mathrm{Ber}\bigl(h_j\mid\sigmoid(\tau_j(\bm{m}_v^{(\mu)}\cup\mathbf{d}^{(\mu)}))\bigr),
\end{align*}
where $\bm{m}_v^{(\mu)} := \{m_{i}^{(\mu)}\mid i\in M^{(\mu)}\}$ and $\bm{m}_h^{(\mu)} := \{m_{j}^{(\mu)}\mid j\in H\}$ are obtained by solving self-inconsistent equations:
\begin{equation}
\begin{split}
m_{i}^{(\mu)} &= \sigmoid(\lambda_i(\bm{m}_h^{(\mu)}))\qquad(i\in M^{(\mu)}), \\
m_{j}^{(\mu)} &= \sigmoid(\tau_j(\bm{m}_{v}^{(\mu)}\cup\mathbf{d}^{(\mu)}))\qquad(j\in H).
\end{split}
\label{eq:mf_equation}
\end{equation}
We solve these equations by randomly initializing $\bm{m}_{v}^{(\mu)}$ and $\bm{m}_{h}^{(\mu)}$ using the uniform distribution $\mathcal{U}(0,1)$ and performing the successive substitution method.
Then, we generate a sample point from the resulting test distribution and use it as an initial point.
This process is repeated $\widehat{K}$ times in parallel to obtain $\widehat{K}$ initial points (see Algorithm~\ref{alg:clamped_initial_points}).
On the clamped distribution, the computational costs of MFA-based initialization and bGS are respectively $O(\widehat{K}|M^{(\mu)}|m)$ and $O(\widehat{K}\widehat{R}|M^{(\mu)}|m)$, where $\widehat{R}$ denotes the number of MCMC steps.
It means that MFA-based initialization does not increase the order of the sampling cost.

\begin{algorithm}
\caption{Proposed generation of initial points for approximating the clamped expectation}
\label{alg:clamped_initial_points}
\begin{algorithmic}[1]
\Require Observed data $\mathbf{d}^{(\mu)}$, missing region $M^{(\mu)}$, sample size $\widehat{K}$
\For{$\nu = 1$ to $\widehat{K}$}
    \State Initialize $m_i^{(\mu)} \sim \mathcal{U}(0,1)$ for $i \in M^{(\mu)}$, and $m_j^{(\mu)} \sim \mathcal{U}(0,1)$ for $j \in H$
    \Repeat 
        \State Update $\bm{m}_v^{(\mu)}$ and $\bm{m}_h^{(\mu)}$ performing the successive substitution method for the equations in Eq.~\eqref{eq:mf_equation}
    \Until{convergence of $\bm{m}_v^{(\mu)}$ and $\bm{m}_h^{(\mu)}$}
    \State $\mathbf{v}_{M^{(\mu)}}^{(\nu)} \sim Q(\bm{v}_{M^{(\mu)}}\mid\mathbf{d}^{(\mu)})$ 
\EndFor
\Ensure Initial points $\bigl\{\mathbf{v}_{M^{(\mu)}}^{(\nu)} \mid \nu = 1, 2, \ldots, \widehat{K} \bigr\}$
\end{algorithmic}
\end{algorithm}

In contrast to Lossy-CD, which generates initial points from a uniform distribution, our method generates them from test distributions obtained via MFA.
Because the test distributions are obtained by minimizing the KLD to the model distribution, the initial points in our proposed method are more likely to closely follow the model distribution than those in Lossy-CD.
However, as mentioned in the literature~\cite{fissore2019robust}, the observed value $\mathbf{v}_{O^{(\mu)}}$ is expected to be correlated with several distinct equilibrium configurations for the missing variable $\bm{v}_{M^{(\mu)}}$ when the missing rate $|M^{(\mu)}|/n$ increases.
In other words, the clamped distribution would be multimodal when the missing rate is high.
This implies that the test distribution may only capture a local peak, rather than the entire target distribution, and thus the initial points generated from these test distributions may not follow the clamped distribution appropriately.
Our experiments show that the performance of the proposed method degrades as the missing rate increases; however, our proposed method still outperforms Lossy-CD.

The free expectation in Eq.~\eqref{eq:free_expect} can be interpreted as a clamped expectation with a missing rate of one, in which case the initial points generated by MFA may not adequately follow the clamped distribution, as discussed above.
Therefore, we refrain from using MFA to generate initial points for approximating the free expectation.
As an alternative, we employ the generation procedure used in standard inverse Ising models, known as PCD~\cite{tieleman2008training}. 
This method reuses the sample points used in the previous parameter update as the initial points and achieves high-quality sampling by bGS with only a few MCMC steps.
These improvements based on MFA or PCD do not increase the computational cost order for sampling, meaning the proposed method allows for sampling at a cost similar to that of Lossy-CD.

\subsection{Improvement on Approximation Based on Spatial Monte Carlo Integration} \label{sec:sec:improvement_approx}

To approximate the clamped and free expectations in Eqs.~\eqref{eq:clamped_expect} and \eqref{eq:free_expect}, we replace the standard MCI used in Lossy-CD with SMCI.
We expand the sum region to include the hidden variables for an accurate approximation.
The sample region is then decided to the visible variables, that is, our SMCI requires sample points only for visible variables.

\begin{figure}
    \centering
    \includegraphics[width=0.95\linewidth]{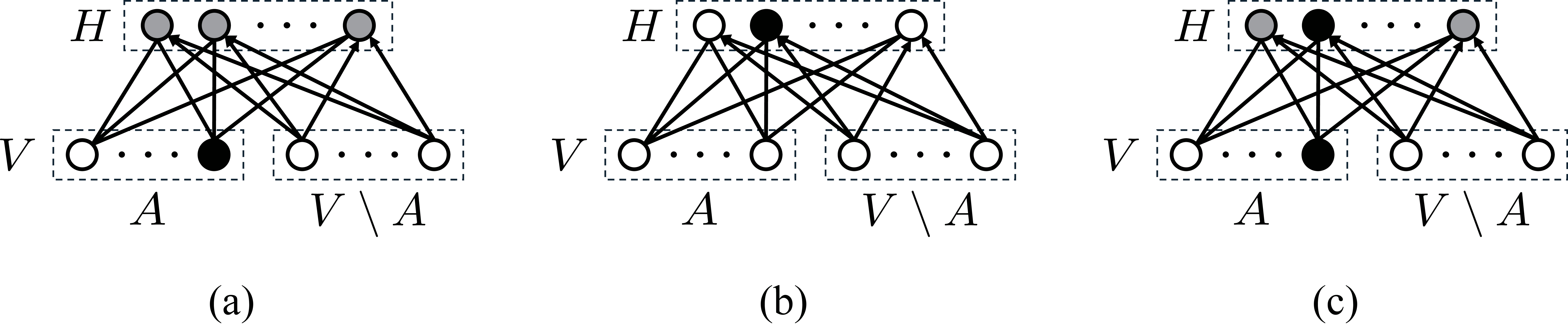}
    \caption{Our sum region settings for the conditional expectation in Eq.~\eqref{eq:cond_expect} under different choices of the function $f(\bm{v}_A,\bm{h})$: (a) $v_i$ for $i\in A$, (b) $h_j$ for $j\in H$, and (c) $v_ih_j$ for $i\in A$ and $j\in H$. The target region is shown in black, and each sum region corresponds to the union of the target region and the region highlighted in gray.}
    \label{fig:smci_on_rbm}
\end{figure}

Consider a conditional expectation expressed as
\begin{align}
\mathbb{E}_\theta[f(\bm{v}_{A},\bm{h})\mid\bm{v}_{V \backslash A}] &= \sum_{\bm{v}_{A}} \sum_{\bm{h}} f(\bm{v}_{A},\bm{h}) P_{\theta}(\bm{v}_{A},\bm{h}\mid \bm{v}_{V \backslash A}).
\label{eq:cond_expect}
\end{align}
This is equivalent to the clamped expectation in Eq.~\eqref{eq:clamped_expect} when $A = M^{(\mu)}$ and the free expectation in Eq.~\eqref{eq:free_expect} when $A=V$.
Thus, the SMCI for Eq.~\eqref{eq:cond_expect} can be applied directly to both cases.
Suppose we obtain $K$ sample points for visible variables drawn from $P_{\theta}(\bm{v}_{A},\bm{h}\mid \bm{v}_{V \backslash A})$, represented as
\begin{align*}
\mathcal{S} := \bigl\{\mathbf{v}_{A}^{(\nu)}\mid\nu=1,2,\ldots,K\bigr\}.
\end{align*}
We approximate the conditional expectation of $f(\bm{v}_A,\bm{h})=v_i$ using 2-SMCI (see Figure~\ref{fig:smci_on_rbm} (a)):
\begin{align}
\mathbb{E}_\theta[v_i\mid\bm{v}_{V\backslash A}] \approx \mathcal{E}_i(\mathcal{S}) &:= \frac{1}{K}\sum_{\nu=1}^K 
\left(\sum_{v_i,\bm{h}} v_i P_\theta(v_i,\bm{h}\mid\bm{v}_{V\backslash A}, \mathbf{v}_{A\backslash\{i\}}^{(\nu)})\right) \nonumber \\
&= \frac{1}{K}\sum_{\nu=1}^K \sigmoid\Bigl[\phi_i\bigl(\bm{v}_{V\backslash A}\cup\mathbf{v}_{A}^{(\nu)}\bigr)\Bigr],  \label{eq:smci_E[v_i]}
\end{align}
where 
\begin{align*}
\phi_i(\bm{v}) &:= b_i + \sum_{j\in H} [\softplus(\tau_{j,i}(\bm{v})+ w_{i,j}) - \softplus(\tau_{j,i}(\bm{v}))], \\
\tau_{j,i}(\bm{v}) &:= \tau_{j}(\bm{v}) - w_{i,j}v_i,
\end{align*}
where $\softplus(x):=\ln(1+e^x)$ denotes the softplus function.
Similarly, we approximate the conditional expectation of $f(\bm{v}_A,\bm{h})=h_j$ using 1-SMCI (see Figure~\ref{fig:smci_on_rbm} (b)):
\begin{align}
\mathbb{E}_\theta[h_j\mid\bm{v}_{V\backslash A}] \approx \mathcal{E}_j(\mathcal{S}) &:= \frac{1}{K}\sum_{\nu=1}^K \left(\sum_{h_j} h_j P_\theta(h_j\mid\bm{v}_{V\backslash A}, \mathbf{v}_{A\backslash\{i\}}^{(\nu)}) \right) \nonumber \\
&= \frac{1}{K}\sum_{\nu=1}^K \sigmoid\Bigl[\tau_i\bigl(\bm{v}_{V\backslash A}\cup\mathbf{v}_{A}^{(\nu)}\bigr)\Bigr],  \label{eq:smci_E[h_j]}
\end{align}
Additionally, we approximate the conditional expectation of $f(\bm{v}_A,\bm{h})=v_ih_j$ by s2-SMCI where the sum region includes the hidden layer (see Figure~\ref{fig:smci_on_rbm} (c)):
\begin{align}
\mathbb{E}_\theta[v_i h_j\mid\bm{v}_{V\backslash A}] \approx \mathcal{E}_{i,j}(\mathcal{S}) &:= \frac{1}{K}\sum_{\nu=1}^K \left(\sum_{v_i,\bm{h}} v_ih_j P_\theta(v_i,\bm{h}\mid\bm{v}_{V\backslash A}, \mathbf{v}_{A\backslash\{i\}}^{(\nu)}) \right) \nonumber \\
&= \frac{1}{K}\sum_{\nu=1}^K \sigmoid\Bigl\{ \logit\bigl[\sigmoid(\tau_{j,i}(\bm{v}_{V\backslash A}\cup\mathbf{v}_{A}^{(\nu)}))\sigmoid(\phi_{i,j}(\bm{v}_{V\backslash A}\cup\mathbf{v}_{A}^{(\nu)}))\bigr] + w_{i,j}\Bigr\},  \label{eq:smci_E[v_ih_j]}
\end{align}
where $\logit(x):=\ln(x / (1-x))$ denotes the logit function known as the inverse function of the sigmoid function, and
\begin{align*}
\phi_{i,j}(\bm{v}) := \phi_{i}(\bm{v}) - [\softplus(\tau_{j,i}(\bm{v})+ w_{i,j}) - \softplus(\tau_{j,i}(\bm{v}))].
\end{align*}

We denote the sample points drawn from the clamped distribution for the $\mu$th data point and the distribution in Eq.~\eqref{eq:RBM_Dist} as $\widehat{\mathcal{S}}^{(\mu)} := \bigl\{\widehat{\mathbf{v}}_{M^{(\mu)}}^{(\nu)}\mid\nu=1,2,\ldots,\widehat{K}\bigr\}$ and $\widetilde{\mathcal{S}} := \bigl\{\widetilde{\mathbf{v}}^{(\nu)}\mid\nu=1,2,\ldots,\widetilde{K}\bigr\}$, respectively.
Using Eqs.~\eqref{eq:smci_E[v_i]}--\eqref{eq:smci_E[v_ih_j]}, the gradients in Eqs.~\eqref{eq:grad_b}--\eqref{eq:grad_w} can be evaluated as
\begin{align*}
\frac{\partial\,\ell_{\mathcal{D}}}{\partial\,b_i} &\approx \frac{1}{N}\sum_{\mu=1}^N \Bigl[I_{O^{(\mu)}}(i)\, \mathrm{d}_i^{(\mu)} + (1 - I_{O^{(\mu)}}(i)) \, \mathcal{E}_i\bigl(\widehat{\mathcal{S}}^{(\mu)}\bigr)  \Bigr] - \mathcal{E}_i(\widetilde{\mathcal{S}}), \\ 
\frac{\partial\,\ell_{\mathcal{D}}}{\partial\,c_j} &\approx \frac{1}{N}\sum_{\mu=1}^N \mathcal{E}_j\bigl(\widehat{\mathcal{S}}^{(\mu)}\bigr) - \mathcal{E}_j(\widetilde{\mathcal{S}}), \\
\frac{\partial\,\ell_{\mathcal{D}}}{\partial\,w_{i,j}} &\approx \frac{1}{N}\sum_{\mu=1}^N \Bigl[I_{O^{(\mu)}}(i)\,  \mathrm{d}_i^{(\mu)}\,\mathcal{E}_j\bigl(\widehat{\mathcal{S}}^{(\mu)}\bigr) + (1 - I_{O^{(\mu)}}(i)) \, \mathcal{E}_{i,j}\bigl(\widehat{\mathcal{S}}^{(\mu)}\bigr)\Bigr] - \mathcal{E}_{i,j}(\widetilde{\mathcal{S}}).
\end{align*}

Let us evaluate the computational cost of the estimators in Eqs.~\eqref{eq:smci_E[v_i]}--\eqref{eq:smci_E[v_ih_j]}.
The computational cost of $\{\phi_i\},\{\tau_{i}\},\{\tau_{j,i}\}$, and $\{\phi_{i,j}\}$, which are required for the evaluation, is $O(K|A|m)$.
Using these precomputed values, each estimator can be evaluated in $O(K)$.
Therefore, the total cost of evaluating the estimators is $O(K|A|m)$, which is lower than the cost of generating $\mathcal{S}$ from $P_{\theta}(\bm{v}_{A},\bm{h}\mid \bm{v}_{V \backslash A})$ using bGS, which is $O(KR|A|m)$ with MCMC step $R$.
In summary of the discussion of the computational cost, the cost of generating the initial values and evaluating the estimators is lower than that of bGS; thus, the computational bottleneck in the proposed method remains bGS.
Therefore, the overall computational cost of the proposed method is similar to that of Lossy-CD.

\section{Numerical Experiment} \label{sec:num_exp}

In this section, we demonstrate the performance of the proposed method through numerical experiments.
The incomplet datasets used for performance evaluation were based on binarized MNIST and CalTech 101 Silhouettes.
The MNIST dataset consists of $70,000$ handwritten digit images of $28\times28$ pixels, $50,000$ of which are used for training and $10,000$ for testing.
Each image is grayscale and was binarized using a threshold of $127.5$ for pixel values.
The CalTech 101 Silhouettes dataset consists of $8,671$ shihouettes of object digital images of $28\times 28$ pixels, $6,364$ used for training and $2,307$ for testing.
Pixels in each training data point were randomly masked with a missing probability $p$ and an RBM was trained using the incomplete dataset.
For binarized MNIST, the number of hidden variables was set to $m=100$.
The connection parameters were initialized using Gaussian-type Xavier's initialization~\cite{glorot2010understanding} and the bias parameters were set to zero.
Maximizing the log-likelihood in Eq.~\eqref{eq:log-likelihood} was performed using mini-batch learning with a batch size of $128$ and AdaMax with default settings from the literature ~\cite{Adam2015}.
For approximating the clamed expectations, the sample size and MCMC step of bGS were set to $\widehat{K}=1$ and $\widehat{R}=16$, respectively.
Similarly, for approximating the free expectations, the sample size and MCMC step of bGS were set to $\widetilde{K}=128$ and $\widetilde{R}=16$, respectively.
For CalTech 101 Silhouettes, the number of hidden variables was set to $m=500$.
The initialization of parameters and log-likelihood maximization settings were the same as for MNIST.
The settings for approximating the clamed expectations, the sample size and the MCMC step of bGS were $\widehat{K}=1$ and $\widehat{R}=16$, respectively.
Also, for approximating the free expectations, the sample size and the MCMC step of bGS were set to $\widetilde{K}=128$ and $\widetilde{R}=16$, respectively.
The performance of Lossy-CD and the proposed method was evaluated using two log-likelihoods: one for the complete training dataset before masking and another for the complete testing dataset.
The former corresponds to the log-likelihood of the standard inverse Ising problem, where a higher value indicates that the RBM ideally fits the original training data well, regardless of missing data.
The latter serves as an indicator of generalization performance.
Because these evaluation indicators involve the free energy of the RBM, which is computationally intractable, we approximated the free energy using annealed importance sampling~\cite{neal2001}.

\begin{figure}[t]
\centering
\begin{tabular}{cc}
\begin{minipage}[b]{0.4\linewidth}
\centering
\includegraphics[width=0.9\linewidth]{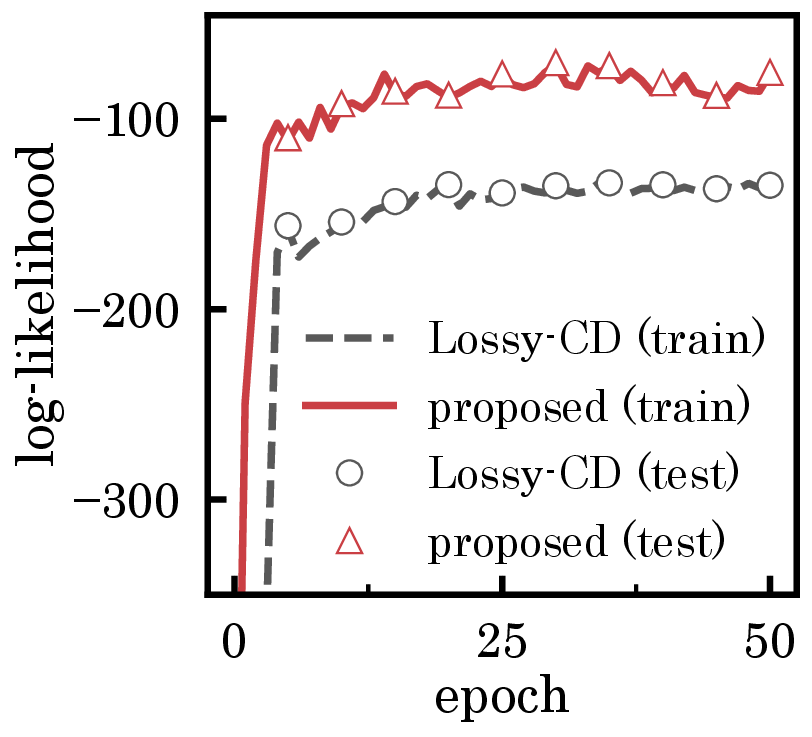} \\
\vspace*{0.1cm}
\hspace*{1.2cm}
(a)
\end{minipage}
\begin{minipage}[b]{0.4\linewidth}
\centering
\includegraphics[width=0.9\linewidth]{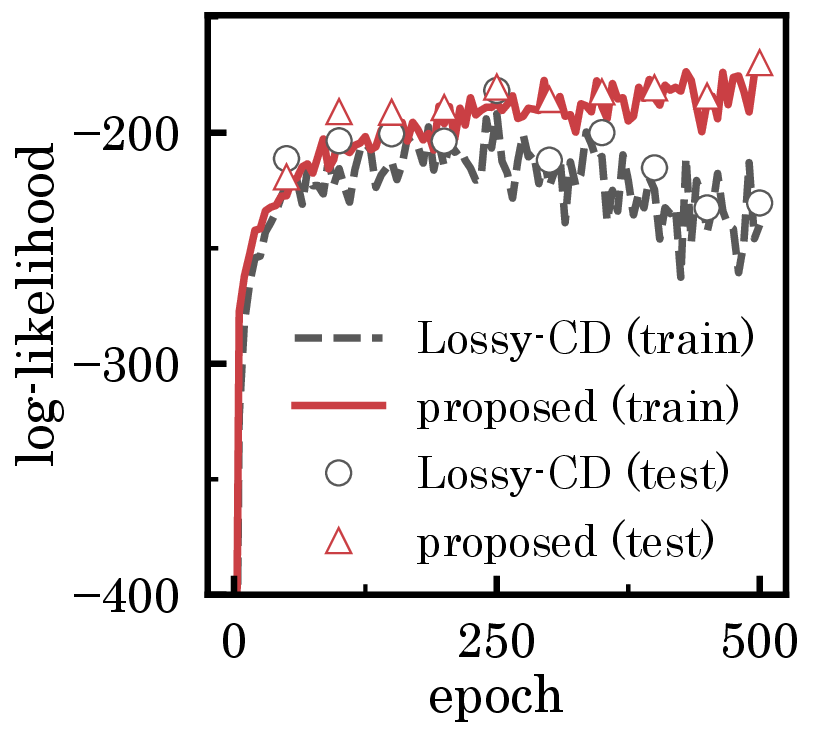} \\
\vspace*{0.1cm}
\hspace*{1.2cm}
(b)
\end{minipage}
\end{tabular}
\caption{Log-likelihoods for the complete training and testing datasets versus epochs in the inverse Ising problem, using (a) binarized MNIST and (b) CalTech 101 Silhouettes datasets with a missing probability of $p=0.3$. Results are averaged over 5 experiments.}
\label{fig:likelihood_vs_epoch}
\end{figure}

\begin{table}[t]
  \centering
  \caption{Log-likelihood evaluations on the complete training and testing datasets of binarized MNIST, using RBMs trained with different missing probabilities $p$. Results are averaged over 5 experiments.}
  \vspace*{2mm}
  \label{tab:mnist_ll}
  \begin{tabular}{lccc|ccc}
    \toprule
    & \multicolumn{3}{c}{Train} & \multicolumn{3}{c}{Test} \\
    \cmidrule(lr){2-4} \cmidrule(lr){5-7}
    Method & $p=0.3$ & $p=0.5$ & $p=0.8$ & $p=0.3$ & $p=0.5$ & $p=0.8$ \\
    \midrule
    Lossy-CD~\cite{fissore2019robust} & $-136$ & $-152$ & $-179$ & $-134$ & $-150$ & $-178$ \\
    Proposed & $-76$  & $-73$  & $-154$ & $-75$  & $-72$  & $-153$ \\
    \bottomrule
  \end{tabular}
\end{table}

\begin{table}[t]
  \centering
  \caption{Log-likelihood evaluations on the complete training and testing datasets of CalTech 101 Silhouettes, using RBMs trained with different missing probabilities $p$. Results are averaged over 5 experiments.}
  \vspace*{2mm}
  \label{tab:caltech_ll}
  \begin{tabular}{lccc|ccc}
    \toprule
    & \multicolumn{3}{c}{Train} & \multicolumn{3}{c}{Test} \\
    \cmidrule(lr){2-4} \cmidrule(lr){5-7}
    Method & $p=0.3$ & $p=0.5$ & $p=0.8$ & $p=0.3$ & $p=0.5$ & $p=0.8$ \\
    \midrule
    Lossy-CD~\cite{fissore2019robust} & $-239$ & $-223$ & $-253$ & $-230$ & $-212$ & $-241$ \\
    Proposed & $-172$ & $-192$ & $-239$ & $-169$ & $-189$ & $-231$ \\
    \bottomrule
  \end{tabular}
\end{table}

Figure~\ref{fig:likelihood_vs_epoch} shows the results of the log-likelihoods against epochs for the binarized MNIST and CalTech 101 Silhouettes datasets.
Here, ``train'' and ``test'' denote the results of the log-likelihoods for the complete training and testing datasets, respectively.
The proposed method consistently achieves faster and higher convergence of the log-likelihoods compared to Lossy-CD.
Tables \ref{tab:mnist_ll} and \ref{tab:caltech_ll} present the log-likelihoods on the complete training and testing datasets for binarized MNIST and CalTech 101 Silhouettes, respectively. 
The results were obtained using RBMs trained with various missing probability values $p$.
The proposed method consistently achieves higher log-likelihood values than Lossy-CD across all conditions.
Although it is evident that the performance of the proposed method degrades as the missing probability increases, it still maintains a superior log-likelihood compared to Lossy-CD, even at higher missing probabilities.

\section{Conclusion} \label{sec:conclusion}

In this study, we proposed an improved method for solving the inverse Ising problem using an incomplete dataset. 
Our approach utilizes an MFA-based initialization for sampling from the clamped distribution, a PCD-based initialization for sampling from the RBM distribution, and SMCI to enhance estimator accuracy. 
These modifications help to accurately address missing data without increasing computational costs.
Experiments on binarized MNIST and CalTech 101 Silhouettes demonstrate that the proposed method converges faster and achieves higher log-likelihoods compared to the conventional Lossy-CD method. 
These results suggest that our approach improves sampling accuracy and provides a robust framework for solving the inverse Ising problem with incomplete datasets.
Future work will focus on applying the proposed method to practical tasks involving RBMs, as outlined in the introduction, to evaluate its real-world efficacy. 
Additionally, we aim to extend our approach to deep Boltzmann machines~\cite{salakhutdinov2009deep} to improve its ability to model complex data distributions.

\bibliographystyle{ieeetr}
\bibliography{reference}

\begin{thebibliography}{10}

\bibitem{smolensky1986}
P.~Smolensky, ``Information processing in dynamical systems: foundations of harmony theory,'' {\em Parallel distributed processing: Explorations in the microstructure of cognition}, vol.~1, pp.~194--281, 1986.

\bibitem{hinton2002}
G.~E. Hinton, ``Training products of experts by minimizing contrastive divergence,'' {\em Neural computation}, vol.~14, no.~8, pp.~1771--1800, 2002.

\bibitem{salakhutdinov2010}
R.~Salakhutdinov and H.~Larochelle, ``Efficient learning of deep boltzmann machines,'' in {\em Proceedings of the thirteenth international conference on artificial intelligence and statistics}, pp.~693--700, JMLR Workshop and Conference Proceedings, 2010.

\bibitem{salakhutdinov2007}
R.~Salakhutdinov, A.~Mnih, and G.~Hinton, ``Restricted boltzmann machines for collaborative filtering,'' in {\em Proceedings of the 24th international conference on Machine learning}, pp.~791--798, 2007.

\bibitem{do2018energy}
K.~Do, T.~Tran, and S.~Venkatesh, ``Energy-based anomaly detection for mixed data,'' {\em Knowledge and Information Systems}, vol.~57, no.~2, pp.~413--435, 2018.

\bibitem{kuremoto2014}
T.~Kuremoto, S.~Kimura, K.~Kobayashi, and M.~Obayashi, ``Time series forecasting using a deep belief network with restricted boltzmann machines,'' {\em Neurocomputing}, vol.~137, pp.~47--56, 2014.

\bibitem{Hatakeyama-Sato2022}
K.~Hatakeyama-Sato, H.~Adachi, M.~Umeki, T.~Kashikawa, K.~Kimura, and K.~Oyaizu, ``Automated design of {Li}+-conducting polymer by quantum-inspired annealing,'' {\em Macromolecular Rapid Communications}, vol.~43, no.~20, p.~2200385, 2022.

\bibitem{torlai2018}
G.~Torlai, G.~Mazzola, J.~Carrasquilla, M.~Troyer, R.~Melko, and G.~Carleo, ``Neural-network quantum state tomography,'' {\em Nature Physics}, vol.~14, no.~5, pp.~447--450, 2018.

\bibitem{carleo2017}
G.~Carleo and M.~Troyer, ``Solving the quantum many-body problem with artificial neural networks,'' {\em Science}, vol.~355, no.~6325, pp.~602--606, 2017.

\bibitem{yasuda2018}
M.~Yasuda, ``Learning algorithm of boltzmann machine based on spatial monte carlo integration method,'' {\em Algorithms}, vol.~11, no.~4, p.~42, 2018.

\bibitem{Sekimoto2023}
K.~Sekimoto and M.~Yasuda, ``Effective learning algorithm for restricted boltzmann machines via spatial monte carlo integration,'' {\em Nonlinear Theory and Its Applications, IEICE}, vol.~14, no.~2, pp.~228--241, 2023.

\bibitem{fissore2019robust}
G.~Fissore, A.~Decelle, C.~Furtlehner, and Y.~Han, ``Robust multi-output learning with highly incomplete data via restricted boltzmann machines,'' {\em arXiv preprint arXiv:1912.09382}, 2019.

\bibitem{tieleman2008training}
T.~Tieleman, ``Training restricted boltzmann machines using approximations to the likelihood gradient,'' in {\em Proceedings of the 25th international conference on Machine learning}, pp.~1064--1071, 2008.

\bibitem{yasuda2015smci}
M.~Yasuda, ``Monte carlo integration using spatial structure of markov random field,'' {\em Journal of the Physical Society of Japan}, vol.~84, no.~3, p.~034001, 2015.

\bibitem{yasuda2021smci}
M.~Yasuda and K.~Uchizawa, ``A generalization of spatial monte carlo integration,'' {\em Neural Computation}, vol.~33, no.~4, pp.~1037--1062, 2021.

\bibitem{doi:10.7566/JPSJ.85.034001}
C.~Takahashi and M.~Yasuda, ``Mean-field inference in gaussian restricted boltzmann machine,'' {\em Journal of the Physical Society of Japan}, vol.~85, no.~3, p.~034001, 2016.

\bibitem{decelle2021restricted}
A.~Decelle and C.~Furtlehner, ``Restricted boltzmann machine: Recent advances and mean-field theory,'' {\em Chinese Physics B}, vol.~30, no.~4, p.~040202, 2021.

\bibitem{neal2001}
R.~M. Neal, ``Annealed importance sampling,'' {\em Statistics and computing}, vol.~11, no.~2, pp.~125--139, 2001.

\bibitem{prat2025mean}
A.~Prat~Pou, E.~Romero, J.~Mart{\'\i}, and F.~Mazzanti, ``Mean field initialization of the annealed importance sampling algorithm for an efficient evaluation of the partition function using restricted boltzmann machines,'' {\em Entropy}, vol.~27, no.~2, p.~171, 2025.

\bibitem{liu2001}
J.~S. Liu, {\em Monte Carlo strategies in scientific computing}.
\newblock Springer, 2001.

\bibitem{glorot2010understanding}
X.~Glorot and Y.~Bengio, ``Understanding the difficulty of training deep feedforward neural networks,'' in {\em Proceedings of the thirteenth international conference on artificial intelligence and statistics}, pp.~249--256, JMLR Workshop and Conference Proceedings, 2010.

\bibitem{Adam2015}
D.~P. Kingma and L.~J. Ba, ``Adam: A method for stochastic optimization,'' {\em In Proc. of the 3rd International Conference on Learning Representations}, pp.~1--13, 2015.

\bibitem{salakhutdinov2009deep}
R.~Salakhutdinov and G.~Hinton, ``Deep boltzmann machines,'' in {\em Artificial intelligence and statistics}, pp.~448--455, PMLR, 2009.

\end{thebibliography}

\end{document}